%% file: main.tex
\documentclass{article}

\PassOptionsToPackage{square,sort,comma,numbers}{natbib}

\usepackage[final]{neurips_2024}

\usepackage[utf8]{inputenc} 
\usepackage[T1]{fontenc}    
\usepackage[hidelinks]{hyperref}        
\usepackage{url}            
\usepackage{booktabs}       
\usepackage{amsfonts}       
\usepackage{nicefrac}       
\usepackage{microtype}      
\usepackage{xcolor}         

\usepackage{graphicx}
\usepackage{subfigure}

\usepackage{algorithm}
\usepackage{algorithmic}

\usepackage{caption}
\usepackage{subcaption}

\usepackage{hyperref}[hidelinks]

\usepackage{amsmath}
\usepackage{amssymb}
\usepackage{mathtools}
\usepackage{amsthm}

\usepackage[capitalize,noabbrev]{cleveref}

\usepackage{url}            
\usepackage{nicefrac}       
\usepackage{microtype}      
\usepackage{xcolor}         
\hypersetup{
    colorlinks,
    linkcolor={black!50!black},
    citecolor={black!50!black},
    urlcolor={blue!80!black}
}

\usepackage{multirow}
\usepackage{amsfonts}
\usepackage{bbm}
\usepackage{arydshln}

\usepackage{float}

\newcommand{\bw}{\mathbf{w}}

\newcommand{\bk}{\mathbf{k}}
\newcommand{\bq}{\mathbf{q}}

\newcommand{\bv}{\mathbf{v}}
\newcommand{\bz}{\mathbf{z}}

\newcommand{\bh}{\mathbf{h}}

\newcommand{\Hcal}{\mathcal{H}}

\newcommand{\argmax}{\mathop{\mathrm{argmax}}\limits}

\theoremstyle{plain}

\theoremstyle{definition}

\theoremstyle{remark}

\title{Decomposable Transformer Point Processes}

\author{%
  Aristeidis Panos \\
  University of Cambridge\\
  \texttt{ap2313@cam.ac.uk} \\
}

\begin{document}

\maketitle

\input{abstract}
\input{introduction}

\input{methodology}

\input{related_work}

\input{experiments}

\input{discussion}
\input{acknowledgements}

\bibliographystyle{plain}
\bibliography{refs}

\input{appendix}
\input{neurips_checklist}

\end{document}

%% file: abstract.tex
\begin{abstract}
The standard paradigm of modeling marked point processes is by parameterizing the intensity function using an attention-based (Transformer-style) architecture. Despite the flexibility of these methods, their inference is based on the computationally intensive thinning algorithm. In this work, we propose a framework where the advantages of the attention-based architecture are maintained and the limitation of the thinning algorithm is circumvented. The framework depends on modeling the conditional distribution of inter-event times with a mixture of log-normals satisfying a Markov property and the conditional probability mass function for the marks with a Transformer-based architecture. The proposed method attains state-of-the-art performance in predicting the next event of a sequence given its history. The experiments also reveal the efficacy of the methods that do not rely on the thinning algorithm during inference over the ones they do. Finally, we test our method on the challenging long-horizon prediction task and find that it outperforms a baseline developed specifically for tackling this task; importantly, inference requires just a fraction of time compared to the thinning-based baseline. 
\end{abstract}

%% file: introduction.tex
\section{Introduction}\label{sec:introduction}
Continuous-time event sequences are commonly found in real-world scenarios and applications such as financial transactions~\cite{bacry2015market}, communication in a social network~\cite{rybski2012communication}, and purchases in e-Commerce systems~\cite{hernandez2017analysis}. This abundance of data for discrete events occuring at irregular intervals has lead to an increasing interest of the community in the last decade to marked temporal point processes which are the standard way of modeling this kind of data.  

Historically, Hawkes processes~\cite{hawkes1971spectra} and Poisson processes~\cite{daley2003basic} have been extensively applied to various domains such as finance~\cite{hasbrouck1991measuring}, seismology~\cite{hawkes1974cluster}, and astronomy~\cite{babu1996spatial}. Despite their elegant mathematical framework and interpretability, the strong assumptions of the models reduce their flexibility and fail to capture the complex dynamics of real-world generating processes. 

Advances in deep learning have allowed the incorporation of neural models like LSTMs~\cite{hochreiter1997long} or recurrent neural networks (RNN) into temporal point processes~\cite{du2016recurrent,mei2017neural,xiao2017modeling,shchur2019intensity,omi2019fully,mei2020neural,boyd2020user}. As a result, these models are able to learn more complex dependencies and attain superior performance than Hawkes/Poisson processes. Recently, the introduction of the (self-) attention mechanism~\cite{vaswani2017attention} to modeling temporal point processes~\cite{zhang2020self,zuo2020transformer,yang2021transformer} has led to new state-of-the-art methods with extra flexibility. 

Despite the advantages of these neural-based models, their dependence on modeling the conditional intensity function creates limitations for both training and inference~\cite{shchur2019intensity}. Training usually requires a Monte Carlo approximation of an integral that appears in the log-likelihood. \cite{omi2019fully} proposed a method to circumvent this approximation; however, the main shortcomings remained as discussed in~\cite{shchur2019intensity}. More importantly, inference is based on the thinning algorithm~\cite{lewis1979simulation,liniger2009multivariate} which is computationally intensive and sensitive to the choice of intensity function. To deal with these downsides,~\cite{shchur2019intensity} parameterized the conditional distribution of the inter-event times by combining a log-normal mixture density network with an RNN. The model's performance is comparable to that of the other intensity-based methods which use RNN/LSTM architecture but still inferior to the Transformer-based methods. 

A more recent work~\cite{panos2023scalable} has referred to the decomposition of the log-likelihood of a marked point process~\cite{cox1975partial} to parameterize the distribution of marks given the time and history and the distribution of times given the history. This decomposition, as with ~\cite{shchur2019intensity}, eliminates the need for the thinning algorithm and additional approximations, while offering a rigorous, yet flexible framework for defining different distributions for occurrence times and marks. \cite{panos2023scalable} used two different parametric models for each distribution, and their results for the time prediction task, despite the simplicity of their framework, were competitive or superior to neural-based baselines.

Inspired by the state-of-the-art performance of the Transformer-based architectures and the computational efficiency/flexibility of the intensity-free models, we develop a model for marked point processes that combines the advantages of these two methodologies. Our contributions are summarized below:
\begin{itemize}
    \item We propose a novel model that is defined by two distributions: a distribution for the marks based on a Transformer architecture and a simple log-normal mixture model for the inter-event times which satisfies a simple Markov property.
    \item Through an extensive experimental study, we show the efficiency of our model in the next-event prediction task  and the suitability of the intensity-free models for correctly predicting the next occurrence time over the  methods relied on the thinning algorithm. 
    \item To the best of our knowledge, we are the first to experimentally show the limitations of the thinning algorithm on the predictive ability of the neural point processes.
    \item We test our model on the more challenging long-horizon prediction task and we provide strong evidence that we can achieve better results in a fraction of time compared to models that have been specifically designed to solve this task and, uncoincidentally, depend on the thinning algorithm.
\end{itemize}

%% file: methodology.tex
\section{Background}\label{sec:background}
A marked temporal point process (MTPP), observed in the interval $(0, T)$, is a stochastic process whose realizations are sequences of discrete events occurring at times $0 < t_1 < \ldots < t_N < T$ with corresponding event types (or marks) $k_1, \ldots, k_N$, where $k_i \in \{1, \ldots, K \}$. The entire sequence is denoted by $\Hcal_T = \{ (t_1, k_1), \ldots, (t_N, k_N)  \}$. The process is fully specified by the conditional intensity function (CIF) of the event of type $k$ at time $t$ conditioned on the event history $\Hcal_{t_i}  = \{(t_j, k_j) \mid t_j < t_i \}$, $\lambda^{\ast}_{k} (t) :=  \lambda_k (t \mid \Hcal_{t_i}) \geq 0, t > t_i$; we use the asterisk $\ast$ to denote the dependence on $\Hcal_{t_i}$. The CIF is used to compute the infinitesimal probability of event $k$ occurring at time $t$, i.e. $\lambda^{\ast}_{k} (t) dt = \mathbb{P}\left(  t_{i+1} \in [t, t+dt], k_{i+1} = k \mid t_{i+1} \notin (t_i, t), \Hcal_{t_i}  \right)$.
 The log-likelihood of such an autoregressive multivariate point process is given by~\cite{hawkes1971spectra,liniger2009multivariate}
\begin{equation}\label{eq:log_lkl_intensity}
  \mathcal{L}(\Hcal_T) = \sum_{i=1}^N \lambda^{\ast}_{k_i} (t_i) - \sum_{k=1}^K \int_0^T \lambda^{\ast}_{k} (t)~dt. 
\end{equation}
Modeling the intensity function by a flexible model and then learning its parameters by maximizing Eq. \eqref{eq:log_lkl_intensity} has been the standard approach of many works~\cite{du2016recurrent,mei2017neural,xiao2017modeling,shchur2019intensity,omi2019fully,mei2020neural,boyd2020user,zhang2020self,zuo2020transformer,yang2021transformer}.

An equivalent way of deriving the log-likelihood in \eqref{eq:log_lkl_intensity} without the use of  $\lambda^{\ast}_{k} (t)$ is by following the decomposition of a multivariate distribution function in~\cite{cox1975partial} (expression 2), expressed as
\begin{equation}\label{eq:log_likelihood_density}
   \mathcal{L}(\Hcal_T) = \sum_{i=1}^N \left\{ \log p^\ast(k_i \mid t_i) + \log f^\ast(t_i) \right\} 
    + \log \left( 1 - F(T \mid \Hcal_{t_N}) \right),
\end{equation}
where $p^\ast(k \mid t_i) :=  p(k \mid t_i, \Hcal_{t_i})$\footnote{The notation of $^\ast$ is slightly different compared to $\lambda^{\ast}_{k}$ but the definition remains consistent.} and $f^\ast(t) := f(t \mid \Hcal_{t_i})$ are the conditional probability mass function (CPMF) of the event types and the conditional probability density function (CPDF) for the occurrence times, respectively.  $F(t \mid \Hcal_{t_i}) = \int_{t_i}^t f^\ast(t) dt~, \forall t > t_i$ is the cumulative distribution function of $f^\ast(t)$. The last term in~\eqref{eq:log_likelihood_density} is the logarithm of the survival function that expresses the probability that no event occurs in the interval $(t_N, T)$. The relation between $\lambda^{\ast}_{k} (t)$ and the density/PMF is given by $\lambda^{\ast}_{k} (t) = \frac{ f^{\ast}(t) p^{\ast}(k \mid t)}{1 - F(t \mid \Hcal_{t})}$; see Section 2.4 in \cite{rasmussen2018lecture}. 

We can represent the temporal part of the process in terms of the inter-event times $\tau_i := t_i - t_{i-1} \in \mathbb{R}_+, t_0 = 0$; the two representations are isomorphic and the relation between the conditional PDF of the inter-event time $\tau_i$ until the next event and the conditional intensity function is given by $g^{\ast}(\tau_i) := g^{\ast}(\tau_i \mid \Hcal_{t_i}) = \sum_{k=1}^K \lambda^{\ast}_{k} (t_{i-1} + \tau_i) \exp \left( - \sum_{k=1}^K \int_0^{\tau_i} \lambda^{\ast}_{k} (t_{i-1} + x) d x \right) = f^\ast(t_i)$.

\section{Decomposable Transformer Point Processes}\label{sec:model}
To develop our proposed framework \emph{Decomposable Transformer Point Process (DTPP)}, we adopt the decomposition in \eqref{eq:log_likelihood_density} and model $g^{\ast}(\tau)$ and $p^\ast(k \mid t)$, separately. Despite the advantages of modeling the intensity function and the arguments in favor of this~\cite{de2019temporal}, we believe that modeling the probability density/mass function offers not only the same benefits as modeling the intensity function as discussed in~\cite{shchur2019intensity}, but, more importantly, it allows us not to depend on the thinning algorithm during inference. The technical details of each model are described in the next two sections.

\subsection{Distribution of Marks}\label{sec:mark_distribution}
The conditional distribution of the event types is parameterized by a continuous-time Transformer architecture as the one described in \cite{yang2021transformer}. More specifically, for any pair of events $(t,k)$, we evaluate an embedding $\bh_k (t) \in \mathbb{R}^D$ based on the history $\Hcal_t$. Assuming an $L$-layer architecture,   $\bh_k (t)$ is given by the concatenation of the embedding of each individual layer, i.e. $\bh_k (t) = [\bh_k^{(0)} (t); \bh_k^{(1)} (t); \ldots; \bh_k^{(L)} (t)]$. The embedding of the base layer $\bh_k^{(0)} (t)$ is independent of time and it is learned by a simple weight vector for each mark, i.e. $\bh_k^{(0)} (t) := \bh_k^{(0)} \in \mathbb{R}^{D^{(0)}}$. The embedding of layer $\ell \in \{1, \ldots, L \}$ for $(t,k)$ is defined as 
\begin{equation}
    \bh_k^{(\ell)} (t)  :=  \bh_k^{(\ell - 1)} (t) + \tanh \left(  \sum_{(t_i, k_i) \in \Hcal_t} \frac{\bv^{(\ell)}_{k_i} (t_i)~\alpha^{(\ell)}_{k_i} (t_i; t, k) }{1 + C}  \right) \in \mathbb{R}^{D^{(\ell)}},
\end{equation}
where $C > 0$ is the normalization constant given by $C =  \sum_{(t_i, k_i) \in \Hcal_t} \alpha^{(\ell)}_{k_i} (t_i, t, k)$ and the unnormalized attention weight is
\begin{equation}
    \alpha^{(\ell)}_{k_i} (t_i; t, k) = \exp \left( \frac{1}{\sqrt{D}} \bk_{k_i}^{(\ell)} (t_i)^{\top} \bq_{k}^{(\ell)} (t) \right) > 0.
\end{equation}
The operation of the non-linear activation function $\tanh$ is element-wise and $D = \sum_{\ell=0}^L D^{(\ell)} $. The query, key, and value  vectors $\bq_{k}^{(\ell)} (t), \bk_{k}^{(\ell)} (t)$, and $\bv_{k}^{(\ell)} (t)$, respectively, can be computed by using the embedding of the previous layer and the corresponding weight matrices $Q^{(\ell)}, K^{(\ell)} \in \mathbb{R}^{D \times (D + D^{(\ell-1)})}, V^{(\ell)} \in \mathbb{R}^{D^{(\ell)} \times (D + D^{(\ell-1)})}$ as follows,
\begin{equation}
  \bq_{k}^{(\ell)} (t)  = Q^{(\ell)} X_t^\ell,~~
  \bk_{k}^{(\ell)} (t)  = K^{(\ell)} X_t^\ell,~~
  \bv_{k}^{(\ell)} (t)  = V^{(\ell)} X_t^\ell \label{eq:qkv},
\end{equation}
where $X_t^\ell =  \left[ \bz(t); \bh_k^{(\ell - 1)} (t) \right] \in  \mathbb{R}^{D + D^{(\ell-1)}}$. By $\bz(t) \in \mathbb{R}^D$, we denote a temporal embedding of time defined as
\begin{align}
 [\bz(t)]_d   = \left\{
\begin{array}{ll}
      \cos \left(t / 10^{\frac{4 (d-1)}{D}} \right),~~\text{if d is odd,} \\
      \sin \left(t / 10^{\frac{4 d}{D}} \right),~~\text{if d is odd,} \\
\end{array} 
\right. 
\end{align}
where $d=0, \ldots, D-1$. This encoding is the same as in \cite{zuo2020transformer} with small differences than the one used in \cite{yang2021transformer} where we found empirically the former to work slightly better than the latter. For a more detailed discussion regarding the  architecture of the model and how it compares to previous Transformer-based methods, see Appendix A in \cite{yang2021transformer}. Finally, we note that for extra model flexibility, multi-head self-attention can be easily obtained by the three equations in \eqref{eq:qkv}.

Having computed the top-layer embeddings $\bh_k (t)$ for all $k=1, \ldots, K$, we model the conditional PMF $p^\ast(k \mid t)$  as
\begin{equation}\label{eq:marks_model}
    p^\ast(k \mid t) = \frac{\exp \left(  \bw_k^{\top}  \bh_k (t)  \right)}{ \sum_{l=1}^K \exp \left(  \bw_l^{\top}  \bh_l (t)  \right)} ,
\end{equation}
where $\bw_k$ are the learnable classifier weights. As is typical for these architectures, to avoid any data leakage from future events, we mask all future events $(t_i, k_i)$ where $t < t_i$ and only use previous events for computing these embeddings.

\subsection{Distribution of Inter-Event Times}\label{sec:time_distribution}
For the modeling of the inter-event times, since they always take positive values, we choose a mixture of log-normal distributions whose parameters depend on the value of the previously seen mark. Specifically, given that the previous occurred mark is $k$, the PDF\footnote{To avoid notation cluttering, we use $\tau$ and $k$ in lieu of  $\tau_i$ and $k_{i-1}$, respectively, to define the PDF.} of the next inter-event time $\tau$ is defined as
\begin{equation}\label{eq:times_model}
     g^{\ast}(\tau) = g(\tau \mid k) = 
     \sum_{m=1}^M w_m^{(k)} \frac{1}{\tau s_m^{(k)} \sqrt{2 \pi}} \exp \left( - \frac{1}{2} \left( \frac{ \log \tau - \mu_m^{(k)}}{s_m^{(k)}} \right)^2  \right),
\end{equation}
where $\{ w_m^{(k)} \}_{m=1}^M  \in \Delta^M$ are the mixture weights, $\{ \mu_m^{(k)} \}_{m=1}^M  \in \mathbb{R}^M$ are the mixture means, and $\{ s_m^{(k)} \}_{m=1}^M \in \mathbb{R}^M_+$ are the standard deviations, for any $k=1, \ldots, K$. The log-normal mixture has several desirable features that justifies our choice: (i) it efficiently approximates distributions in low dimensions such as 1-d distributions of inter-event times~\cite{mclachlan2019finite,shchur2019intensity} while satisfying a universal approximation property that provides theoretical guarantees regarding its approximation ability~\cite{dasgupta2008asymptotic}, (ii) closed-form moments are available and can be used for predicting the next time; for instance, the mean of the distribution is given as the weighted average of each of the log-normal means, i.e. 
\begin{equation}
\mathbb{E}^{(k)}_g[\tau] =  \sum_{m=1}^M w_m^{(k)} \exp \left( \mu_m^{(k)} +  \frac{(s_m^{(k)})^2}{2}  \right),  
\end{equation}
(iii) learning the small number of parameters $\{ w_m^{(k)}, \mu_m^{(k)}, s_m^{(k)} \}_{m=1}^M$ can be done in a fraction of time using fast off-the-shelf implementations based on the EM algorithm~\cite{dempster1977maximum}. Finally, note that the dependence of the model only on the most recent mark implies a Markov property since we do not need the entire history $\Hcal_{<t}$ to define our distribution. 

At first glance, this assumption might seem restrictive when it comes to capturing the complex dynamics of the process. Nevertheless, this assumption holds only for $g^\ast$ while $p^\ast$ is modeled by the flexible Transformer architecture that models the full history up to the current time $t$. Hence, we do not sacrifice any modeling power at all to achieve efficiency. On the contrary, we can maintain both modeling power and computational efficiency due to the decomposition in \eqref{eq:log_likelihood_density} and the chosen models in \eqref{eq:times_model} and \eqref{eq:marks_model}. Moreover, as our extensive experiments on the real-world data show, this assumption provides a robust predictive model which is less prone to overfitting compared to more flexible neural net architectures since the Markov property can act as a strong regularizer.

\subsection{Training and Prediction}
The parameters $\{ w_m^{(k)}, \mu_m^{(k)}, s_m^{(k)} \}_{m, k}$, of $g^{\ast}(\tau)$ and the parameters $\{\bw_k, \bh_k^{(0)}, Q^{(\ell)}, K^{(\ell)}, V^{(\ell)} \}_{\ell, k}$ of $p^\ast(k \mid t)$ can be estimated by maximizing the log-likelihood in \eqref{eq:log_likelihood_density} using any stochastic gradient method. A crucial benefit of using the decomposition in \eqref{eq:log_likelihood_density} is that it permits us to learn the above parameters separately as follows:
\begin{align}
\{ w_m^{(k)^\ast}, \mu_m^{(k)^\ast}, s_m^{(k)^\ast} \}_{m, k}   & =  \argmax  \sum_{i=1}^N \log g^\ast(\tau_i) + \log \left( 1 - G(T \mid \Hcal_{T}) \right) \label{eq:mle_times} \\
\{\bw^{\ast}_k, \bh_k^{(0)^\ast}, Q^{(\ell)^\ast}, K^{(\ell)^\ast}, V^{(\ell)^\ast} \}_{\ell, k}  & = \argmax \sum_{i=1}^N \log p^\ast(k_i \mid t_i),  \label{eq:mle_marks}
\end{align}
where $G(T \mid \Hcal_{T} ) = \int_{0}^{T-t_N} g^\ast(\tau) d \tau~$. This is a major difference from previous work where the parameters of the neural net parameterizing the conditional intensity function had to be learned all at once. By dividing the main objective function into two sub-objectives, since there is no parameter-sharing between the two models, we can maximize the two objectives independently, and thus, having an easier optimization task than maximizing a single set of parameters of a given objective, such as \eqref{eq:log_lkl_intensity}.

The trained models $g^\ast$ and $p^\ast$ can now be used to predict either the time/type of the next event (next-event prediction) or the next $P > 1$ events (long-horizon prediction). For the next-event-prediction, the predicted time $\hat{t}$ given the history $\Hcal_{t_i}$ is computed by using the mean of the appropriate mixture of log-normals while the corresponding predicted type of this event $\hat{k}$ is evaluated based on $\Hcal_{t_i}$ and the true time $t_{i+1}$, i.e.
\begin{equation}
 \hat{t}  = t_i + \mathbb{E}^{(k_i)}_g[\tau],~~~~~~\hat{k}  = \argmax_k p^\ast(k \mid t_{i+1}).\label{eq:predict_next_event}
\end{equation}
The above procedure is based on the minimum Bayes risk (MBR) principle~\cite{mei2017neural} which aims to predict the time and type that minimizes the expected loss. This is an average $L_2$ loss in the case of time prediction (deriving a root mean squared error) and an average 0-1 loss for the type prediction (deriving an error rate).

For the long-horizon prediction task~\cite{deshpande2021long,xue2022hypro}, we need to predict a sequence of events where, unlike the next-event prediction task, we do not have access to the true time when we predict the next event type. This could potentially lead to a cascading error effect due to the autoregressive nature of the models designed for the less challenging task of next-event prediction. This is because after an error is made in the sequence of the predictions, it cannot be corrected, and thus the error accumulates and affects all subsequent predictions. We argue that this pathology can be alleviated by using a model of times that provides accurate and robust predictions given the history. This assumption is verified experimentally in Section~\ref{sec:long_hor_pred}. To generate a predicted sequence, we require the trained models $g()$ and $p()$ to sequentially predict events as in \eqref{eq:predict_next_event}. Since we have no access to the true time $t_{i+1}$, we use as a proxy the prediction $\hat{t}$ to predict $\hat{k}$ in turn. After the prediction of the new event, we append it to the history and then we repeat the same step given the updated history until we generate a sequence of $P$ events. The exact procedure is described in Algorithm~\ref{alg:long_horizon} in the Appendix.  

The main advantage of Algorithm~\ref{alg:long_horizon} over other methods~\cite{xue2022hypro} that are based on the thinning algorithm is its computational efficiency. The algorithm is fully parallelizable, and it can produce single steps in parallel for a batch of event sequences. This is not possible for thinning-based methods that require one to consider single sequences each time~\cite{xue2022hypro}. Consequently, our method is able to generate sequences orders of magnitude faster, which is verified by our experiments. Unlike other competitors~\cite{xue2022hypro} that are based on the thinning algorithm and therefore require random sampling, our algorithm is fully deterministic; for comparison the thinning algorithm is described in Algorithm~\ref{alg:thinning} of the Appendix.

\begin{figure} 
    \begin{center}
     \subfigure[Mimic-II]{%
        \includegraphics[width=0.3\linewidth, height=0.25\textwidth]{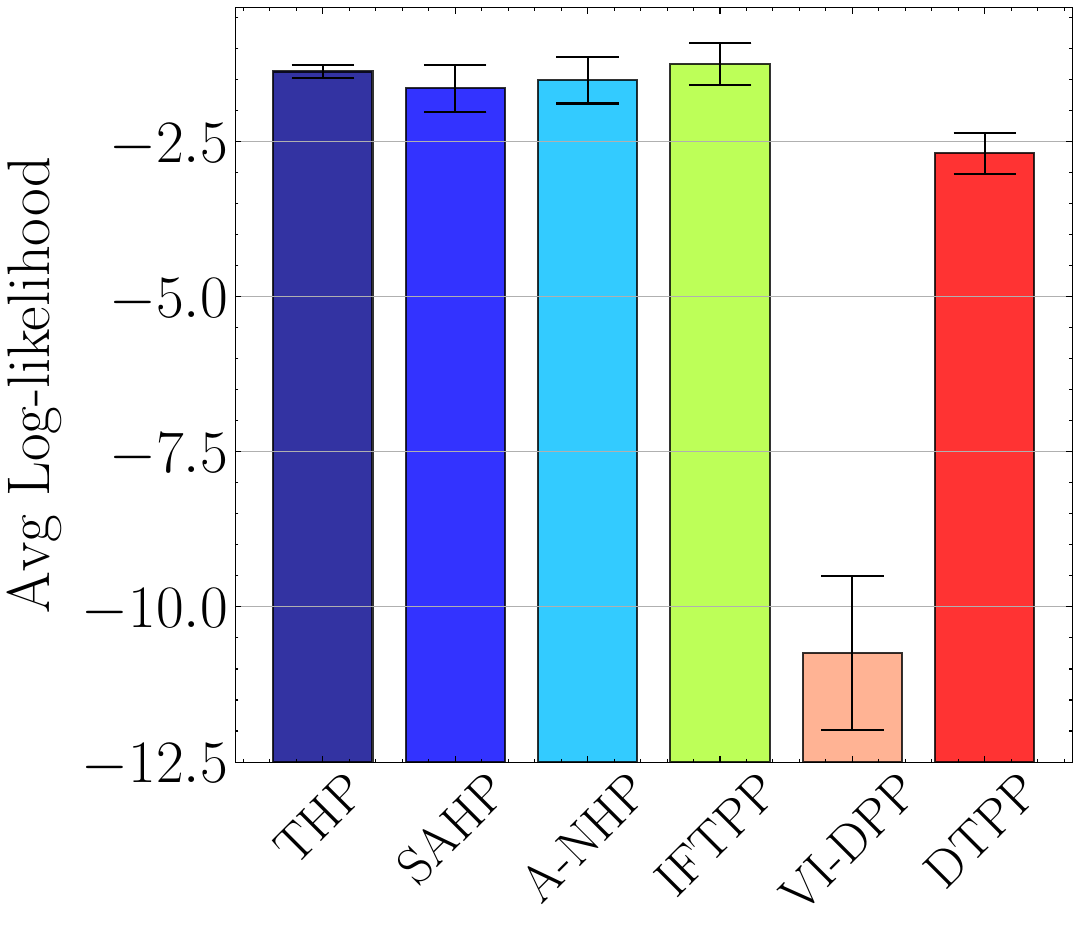}
        \label{fig:mimic_log_lkl}}
     \subfigure[Amazon]{
        \includegraphics[width=0.3\linewidth, height=0.25\textwidth]{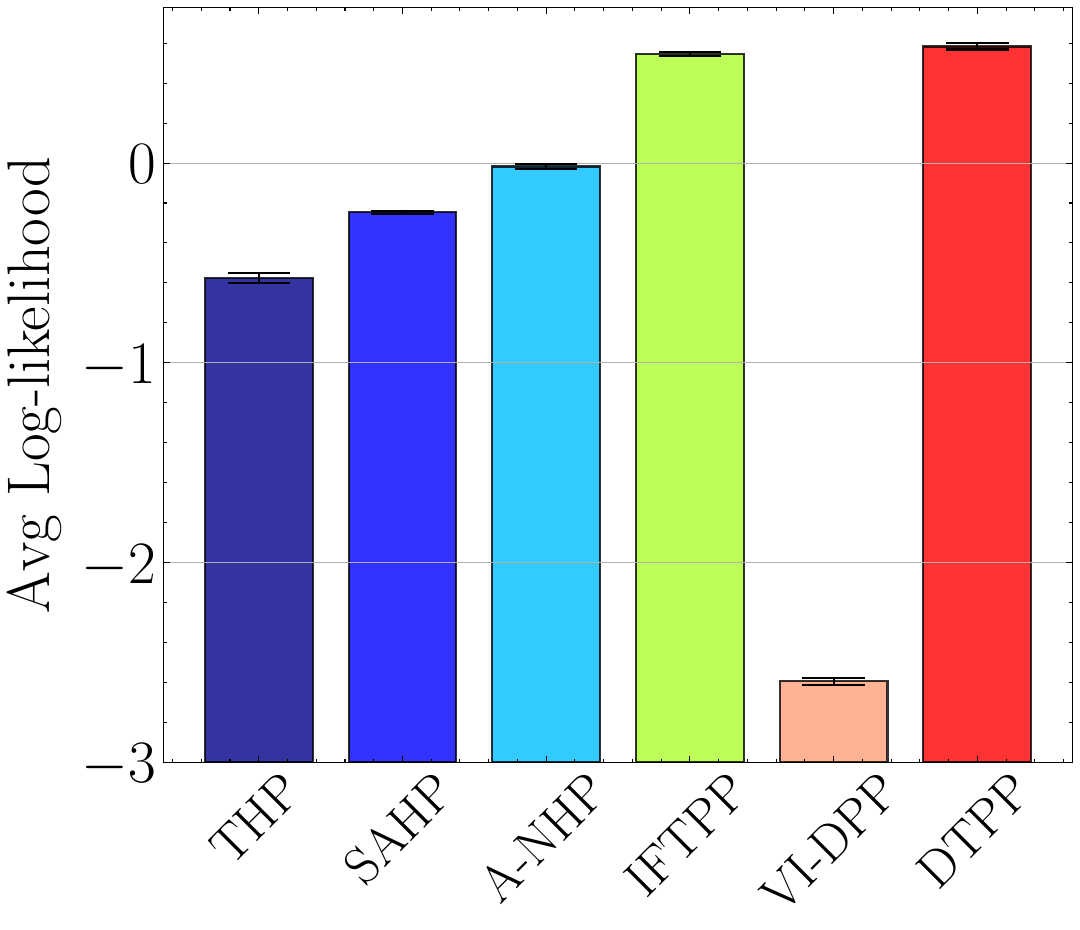}
        \label{fig:amazon_log_lkl}
        }
         \subfigure[Taxi]{
        \includegraphics[width=0.3\linewidth, height=0.25\textwidth]{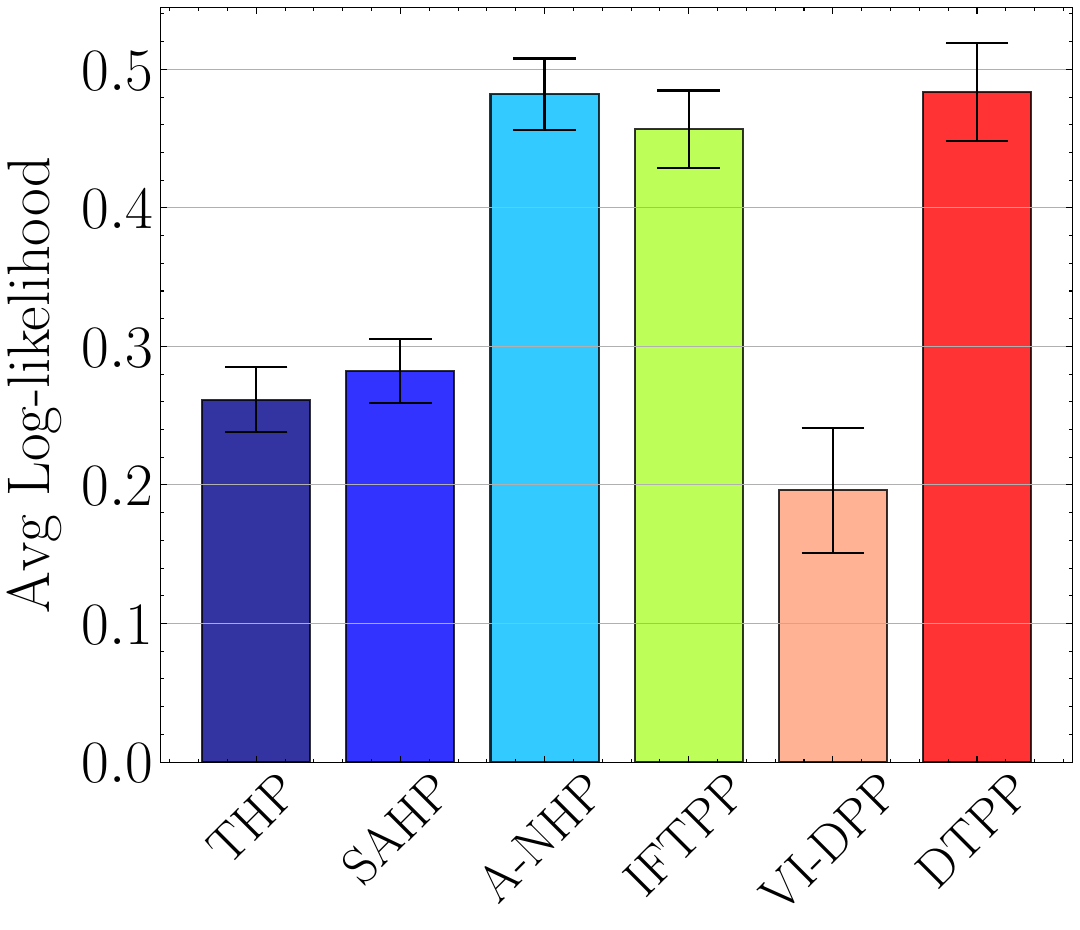}
        \label{fig:taxi_log_lkl}
        }
     \subfigure[Taobao]{
        \includegraphics[width=0.3\linewidth, height=0.25\textwidth]{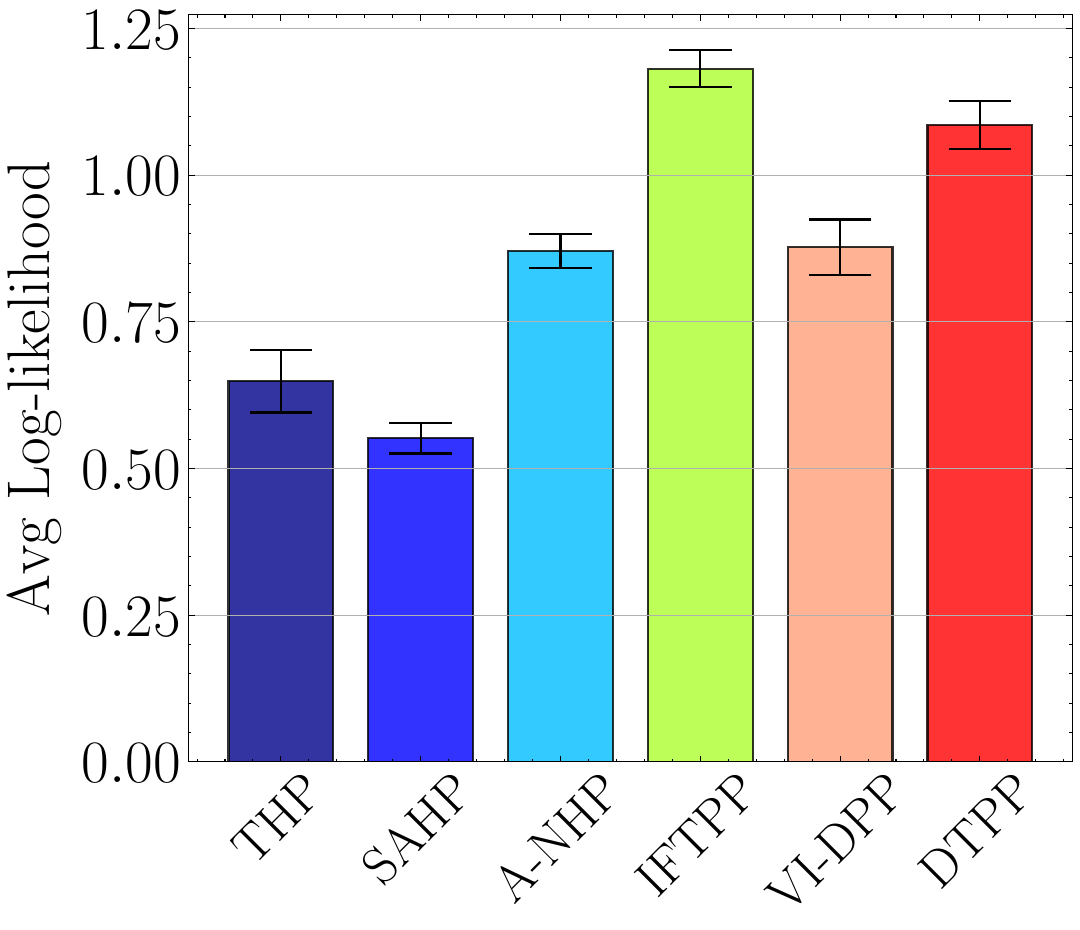}
        \label{fig:taobao_log_lkl}
        }
    \subfigure[StackOverflow]{
        \includegraphics[width=0.3\linewidth, height=0.25\textwidth]{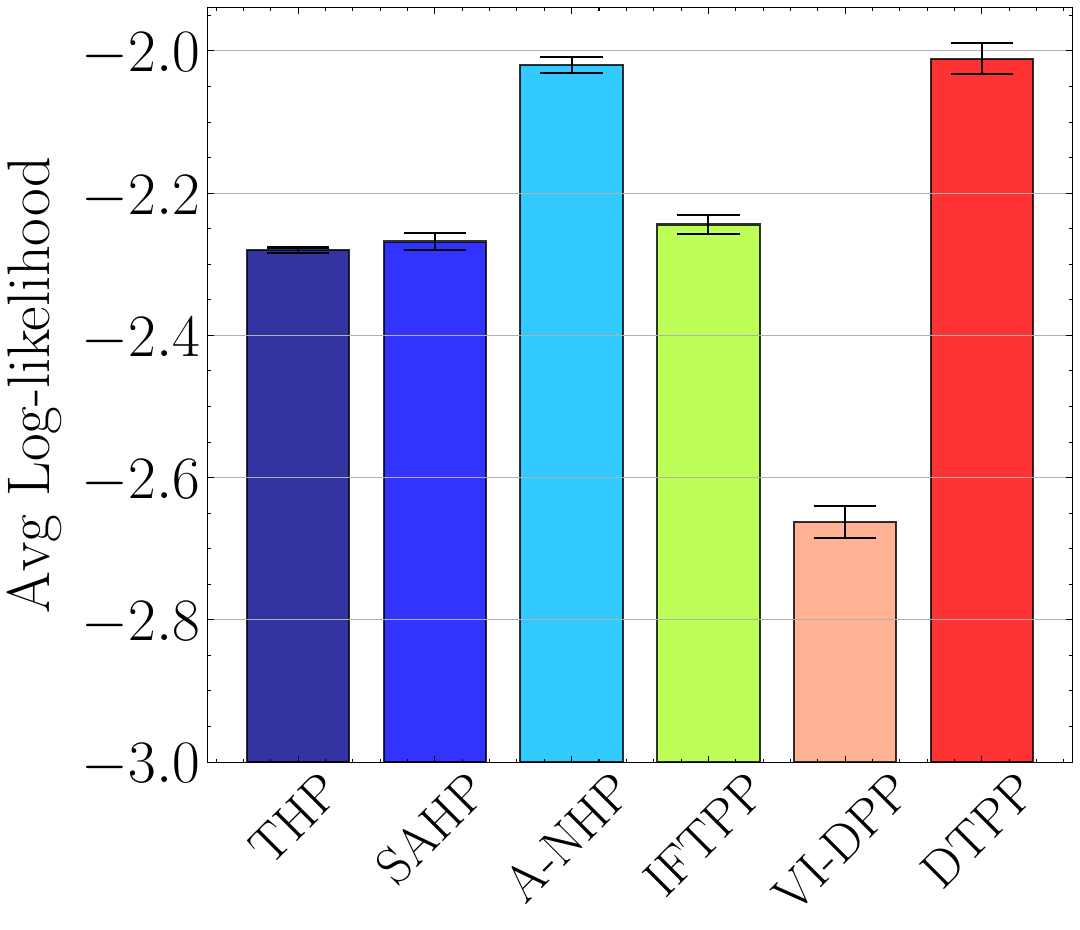}
        \label{fig:so_log_lkl}
        }
    \caption{Goodness-of-fit evaluation over the five real-world datasets. We compare our DTPP model against five strong baselines. Results (larger is better) are accompanied by 95\% bootstrap confidence intervals.}\label{fig:log_lkl}
    \end{center}
\end{figure}

%% file: related_work.tex
\section{Related Work}\label{sec:related_work}
The decomposition in~\eqref{eq:log_likelihood_density} has been used in the past to provide both expressive and interpretable models~\cite{narayanan2023flexible,panos2023scalable}. For instance, \cite{panos2023scalable} model the mark distribution with a parametric model, inspired by the exponential intensity function of a Hawkes process and the time distribution with a single log-normal; however, they use the mode instead of the mean of the distribution for predicting the time of the next event. They learn the parameters of their models separately, as we describe in~\eqref{eq:mle_times} and~\eqref{eq:mle_marks}, but they use Variational Inference~\cite{blei2017variational} to learn the parameters of $p^\ast$. They attain competitive results in terms of next-time prediction, but the model lacks the flexibility of a Transformer-based architecture, as our experiments show. 

\cite{shchur2019intensity} is another work that takes advantage of the mixture of log-normal distributions to model the distribution of the inter-event times. The model is based on an RNN architecture that produces a fixed-dimensional embedding of the event history, which is used to generate the parameters of the mixture model, and the same embedding is employed to define the CPMF of the marks. In our case, we use a Transformer architecture to obtain this history embedding, which is utilized by the CPMF, exclusively. Finally, the proposed model in \cite{shchur2019intensity} assumes that the marks are conditionally independent of the time given the history, which is not the case for our framework, as is evident in~\eqref{eq:log_likelihood_density}. 

Finally, the CPMF of the marks for our DTPP model shares the same architecture as the Attentive Neural Hawkes Process (A-NHP)~\cite{yang2021transformer}. Nevertheless, they use it to model the CIF while in our case we model $p^\ast$.

%% file: experiments.tex
\section{Experiments}\label{sec:experiments}
We considered two different tasks to assess the predictive performance of our proposed method: Goodness-of-fit/next-event prediction and long-horizon prediction. We compared our method DTPP to several strong baselines over five real-world datasets and three synthetic ones.  Description and summary statistics for all datasets used in this section are given in Appendix~\ref{sec:dataset_details}. For the competing methods, we used their published implementations; more details are given in \ref{sec:implementation_details}. Experimental details not available in this section can be found in Appendix \ref{sec:exp_details}. Our framework was implemented with PyTorch \cite{paszke2019pytorch} and scikit-learn \cite{scikit-learn}; the code is available at \url{https://github.com/aresPanos/dtpp}.

\subsection{Goodness-of-Fit / Next-Event Prediction}\label{sec:next_event_pred}
We evaluated our \textbf{DTPP} model to determine how well it generalizes and predicts the next event given the history on the held-out dataset. For comparison, we used five state-of-the-art baselines where the three of them model the CIF using Transformers, while the other two model the CPDF of inter-event times and the CPMF of marks (see Section~\ref{sec:related_work}). The CIF-based baselines are the \textbf{Transformer Hawkes Process (THP)}~\cite{zuo2020transformer}, the \textbf{Self-Attentive Hawkes Process (SAHP) }~\cite{zhang2020self}, and the \textbf{Attentive Neural Hawkes Process (A-NHP)}~\cite{yang2021transformer}. The CPDF-based ones are the \textbf{Intensity-Free Temporal Point Process (IFTPP) }~\cite{shchur2019intensity}, and the \textbf{VI-Decoupled
Point Process (VI-DPP)}~\cite{panos2023scalable}. 

\begin{table}
\setlength{\tabcolsep}{1.4pt}
\caption{Performance comparison  between our model DTPP and various baselines in terms of next-event prediction. The root mean squared error (RMSE) measures the error of the predicted time of the next event, while the error rate (ERROR-$\%$)  evaluates the error of the predicted mark given the true time. The results (lower is better) are accompanied by 95\% bootstrap confidence intervals. $^\dagger, ^\triangleleft, ^\triangleright$ denote the CIF-based methods, the CPDF-based methods that use a single model, and the ones using a seperate model, respectively.
}\label{table:single_step_pred}

\begin{center}
\begin{small}
\begin{sc}
\begin{tabular}{lcccccccr}
\toprule
& \multicolumn{2}{c}{ \textbf{Amazon}} & \multicolumn{2}{c}{ \textbf{Taxi}} & \multicolumn{2}{c}{ \textbf{Taobao}} & \multicolumn{2}{c}{ \textbf{StackOverflow-V1}} \\
\cmidrule(lr){2-3} \cmidrule(lr){4-5} \cmidrule(lr){6-7} \cmidrule(lr){8-9} \textbf{Methods} & \textbf{RMSE} &  \textbf{Error} &  \textbf{RMSE} &  \textbf{Error} &  \textbf{RMSE} &  \textbf{Error} &  \textbf{RMSE} &  \textbf{Error} \\
\midrule
THP$^\dagger$ & 0.62 {\tiny $\pm 0.03$} & 65.06 {\tiny $\pm 1.04$} & 0.37 {\tiny $\pm 0.02$} & 8.55 {\tiny $\pm 0.65$} & 0.13 {\tiny $\pm 0.02$} & 41.43 {\tiny $\pm 0.78$} & 1.32 {\tiny $\pm 0.03$} & 53.86 {\tiny $\pm 0.76$} \\
SAHP$^\dagger$ & 0.58 {\tiny $\pm 0.01$} & 62.58 {\tiny $\pm 0.32$} & 0.28 {\tiny $\pm 0.04$} & 8.37 {\tiny $\pm 0.43$} & 2.26 {\tiny $\pm 0.59$} & 45.63 {\tiny $\pm 0.58$} & 1.93 {\tiny $\pm 0.04$} & 53.00 {\tiny $\pm 0.32$} \\
A-NHP$^\dagger$ & 0.42 {\tiny $\pm 0.01$} & 65.94 {\tiny $\pm 0.31$} & 0.29 {\tiny $\pm 0.02$} & 7.67 {\tiny $\pm 0.44$} & 1.44 {\tiny $\pm 0.53$} & 43.96 {\tiny $\pm 0.55$} & 1.18 {\tiny $\pm 0.01$} & 52.17 {\tiny $\pm 0.32$} \\
IFTPP$^\triangleleft$ & 0.41 {\tiny $\pm 0.05$} & 64.08 {\tiny $\pm 0.34$} & 0.39 {\tiny $\pm 0.09$} & 8.17 {\tiny $\pm 0.46$} & 0.33 {\tiny $\pm 0.00$} & 41.92 {\tiny $\pm 0.54$} & 1.91 {\tiny $\pm 0.10$} & 53.68 {\tiny $\pm 0.30$} \\
VI-DPP$^\triangleright$ & 0.38 {\tiny $\pm 0.00$} & 65.49 {\tiny $\pm 0.34$} & 0.11 {\tiny $\pm 0.02$} & 9.49 {\tiny $\pm 0.42$} & 0.07 {\tiny $\pm 0.00$} & 41.89 {\tiny $\pm 0.57$} & 1.68 {\tiny $\pm 0.04$} & 55.06 {\tiny $\pm 0.32$} \\
DTPP$^\triangleright$ & \textbf{0.12} {\tiny $\pm 0.00$} & \textbf{59.06} {\tiny $\pm 0.35$} & \textbf{0.08} {\tiny $\pm 0.01$} & \textbf{7.12} {\tiny $\pm 0.40$} & \textbf{0.05} {\tiny $\pm 0.00$} & \textbf{40.12} {\tiny $\pm 0.59$} & \textbf{1.07} {\tiny $\pm 0.03$} & \textbf{50.41} {\tiny $\pm 0.31$} \\
\bottomrule
\end{tabular}
\end{sc}
\end{small}
\end{center}

\end{table}

We fit the above six models on a diverse collection of five popular real-world datasets, each with varied characteristics: \textbf{MIMIC-II}~\cite{lee2011open}, \textbf{Amazon}~\cite{ni2019justifying}, \textbf{Taxi}~\cite{whong2014foiking}, \textbf{Taobao}~\cite{zhu2018learning}, and \textbf{StackOverlfow-V1}~\cite{leskovec2014snap, yang2021transformer}. Training details are given in Appendix~\ref{sec:training_details}.

\paragraph{Goodness-of-Fit.} Figure~\ref{fig:log_lkl} shows the average log-likelihood for each model on the held-out data of the five real-world datasets. Our DTPP model consistently outperforms the simple parametric VI-DPP, indicating the flexibility of using the self-attention mechanism to model the CPDF. Except Mimic-II, DTPP achieves the highest or the second highest log-likelihood across the remaining datasets. Therefore, the separate parameterization of $p^\ast$ and $g^\ast$ does not hurt performance compared to the models with a common set of learnable parameters. Finally, notice that DTPP outperforms all the CPDF-based methods on average, while the two CPDF-based methods that employ deep learning architectures, i.e., DTPP and IFTPP, exhibit better performance than the CIF-based baselines. A plausible explanation is that the log-likelihood computation for the CIF-based baselines requires Monte Carlo integration, which could cause approximation errors; for the CPDF-based methods, this computation is exact. A-NHP is the clear winner among the CIF-based methods, as also shown in~\cite{yang2021transformer}.

\paragraph{Next-Event Prediction.} We evaluate the predictive capacity of all models by predicting each event $(t_i, k_i)$ given its history $\Hcal_{t_i}, i=2, \ldots, N$ on held-out data. Event time prediction is measured by root mean squared Error (RMSE) and event type prediction by error rate; Table~\ref{table:single_step_pred} summarizes the results. DTPP outperforms all the baselines in both tasks. The wider performance gaps in RMSE between our model and the other baselines justify our choice of a inter-event distribution satisfying a Markov property; this result also implies that we do not need long event histories to capture the dynamics of these datasets. We also compare the average performance between CIF-based and CPDF-based (excluding VI-DPP) methods. We see that for the CIF-based baselines the average RMSE is 0.58 and the average error rate is 35.39\%  while  for the CPDF-based ones we have 0.95 and 37.0\%, respectively. These results support our argument that the thinning algorithm tends to harm the time prediction accuracy; they also highlight the efficiency of using a separate model for the inter-event times. Additional results on Mimic-II can be found in Appendix \ref{sec:extra_results_mimic}.

\begin{figure}[ht]
\begin{center}
\centerline{\includegraphics[height=3.9cm]{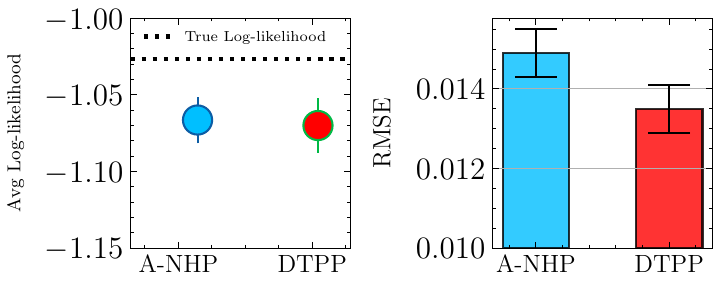}}
\caption{Performance comparison between DTPP and A-NHP over the SAHP-Synthetic dataset.}
\label{fig:sahp_synthetic}
\end{center}
\end{figure}

\paragraph{Synthetic datasets.} To extra investigate the capabilities of our model in a more controlled manner, we created a dataset by generating sequences from a randomly initialized SAHP model. Although, each event has strong dependence on its history, Figure~\ref{fig:sahp_synthetic} shows that our model approximates the true log-likelihood as well as A-NHP. Moreover, DTPP's mixture model is more accurate than the thinning-based A-NHP in time prediction. Moreover, we found  that the only case that DTPP was significantly outperformed by A-NHP was on a synthetic dataset generated by a 1-d Hawkes process. Since no event types are present we only use a single mixture of log-normals which apparently is the wrong model for this data. The results are illustrated in Figure~\ref{fig:hawkes_synthetic} of the Appendix. 

\subsection{Long-Horizon Prediction}\label{sec:long_hor_pred}
To test the performance of our model for this task, we followed the experimental setup of~\cite{xue2022hypro}. From the same work, we used the proposed \textbf{ HYPRO}, which is the state-of-the-art method for the long-horizon prediction task, to compare with DTPP. HYPRO is a globally normalized model that aims to address cascading errors that occur in auto-regressive and locally normalized models, such as the models in Section~\ref{sec:next_event_pred}. HYPRO and DTPP are based on the same Transformer architecture of A-NHP so the main difference is that HYPRO is a CIF-based method, and thus, it requires the thinning algorithm to sample sequences. For our experiments, we use the distance-based regularization variant of HYPRO with a Multi-NCE loss as this method attains the best results in~\cite{xue2022hypro}. As DTPP and HYPRO share the same Transformer architecture, so we used the exact same hyperparameters for fair comparison. Note that even in this case, HYPRO has more than double number of parameters compared to DTPP since HYPRO requires an extra Transformer to model the energy function used for global 
 normalization. More details on HYPRO training and hyperparameters can be found in the Appendix~\ref{sec:exp_details}.

\begin{figure*}[hpbt] 
\begin{center}
         \subfigure[Taxi]{
        \includegraphics[width=0.31\linewidth]{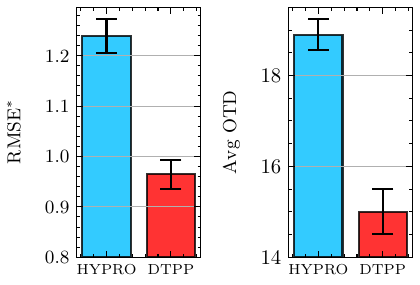}
        \label{fig:taxi_multistep_scores}
        }
     \subfigure[Taobao]{
        \includegraphics[width=0.31\linewidth]{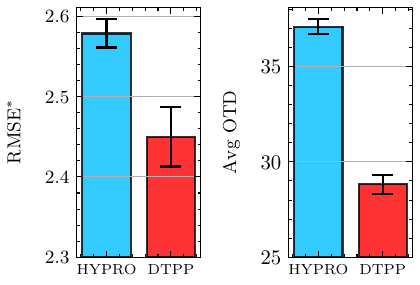}
        \label{fig:taobao_multistep_scores}
        }
    \subfigure[StackOverflow-V2]{
        \includegraphics[width=0.31\linewidth]{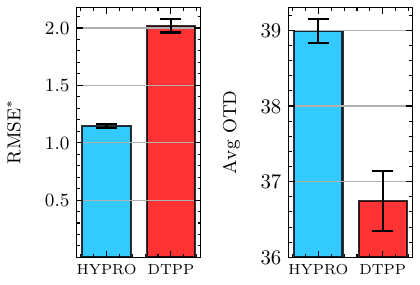}
        \label{fig:so_multistep_scores}
        }
    \caption{Performance comparison over the three real-world datasets measured by RMSE$^\star$ and average OTD (lower is better). The reported results for HYPRO are based on 16 weighted samples, i.e. $M=16$ for Algorithm 2 in~\cite{xue2022hypro}.}\label{fig:multistep_scores}
    \end{center}
\end{figure*}

We used three of the previous real-world datasets for evaluation because of their long sequences. These are \textbf{Taxi}, \textbf{Taobao}, and \textbf{StackOverflow-V2}~\cite{leskovec2014snap,xue2022hypro}. For each dataset, our goal is to predict the last 20 events in a sequence, denoted by $\Hcal_P$, given the history; that is, $P=20$ in Algorithm~\ref{alg:long_horizon}. As is typical for the long-horizon prediction, the standard scores used for evaluating the model's performance are the The optimal transport distance (\textbf{OTD})~\cite{mei2019imputing} and the long-horizon RMSE (\textbf{RMSE$^\ast$)}~\cite{xue2022hypro}.

In Figure~\ref{fig:multistep_scores}, we see that our DTPP method outperforms HYPRO across all datasets in terms of average OTD and RMSE. HYPRO achieves a lower RMSE score only in StackOverflow. These results provide corroborating evidence on our argument that the thinning algorithm might negatively affect the performance of a neural point process even in the case of globally normalized models as HYPRO. It is also evident that a locally normalized CPDF-based model such as DTPP is much more robust against the cascading error which CIF-based methods are vulnerable~\cite{yang2021transformer}. We believe that this robustness stems from the accurate predictions of the log-normal mixture model.

\begin{table*}[ht!]
\setlength{\tabcolsep}{2.9pt}
\caption{Performance comparison between our model DTPP and HYPRO for the long-horizon prediction task. For HYPRO, we use $\{2, 4, 8, 16, 32 \}$ weighted proposals (Algorithm 2 in \cite{xue2022hypro}). We report the average optimal transport distance (avg OTD) and the time (in minutes) required for predicting all the long-horizon sequences of the held-out dataset (lower is better). ``Params'' denotes the number ($\times 10^3$) of trainable parameters of each method. We include error bars based on five runs.  
}
\label{table:multi_step_pred_otd_vs_time}

\begin{center}
\begin{small}
\begin{sc}
\begin{tabular}{lccccccr}
\toprule
& & \multicolumn{2}{c}{ \textbf{Taxi}} & \multicolumn{2}{c}{ \textbf{Taobao}} & \multicolumn{2}{c}{ \textbf{StackOverflow-V2}} \\
\cmidrule(lr){3-4} \cmidrule(lr){5-6} \cmidrule(lr){7-8} \textbf{Methods} &  Params &  \textbf{avg OTD} &  \textbf{Time}  &  \textbf{avg OTD} &  \textbf{Time}  &  \textbf{avg OTD} &  \textbf{Time } \\
\midrule
HYPRO-2 & \multirow{5}{*}{850} & 20.35 {\scriptsize $\pm 0.24$} & 44.81 {\scriptsize $\pm 0.01$} & 39.73 {\scriptsize $\pm 0.37$} & 43.53 {\scriptsize $\pm 0.03$} & 39.84 {\scriptsize $\pm 0.12$} & 46.32 {\scriptsize $\pm 0.01$} \\
HYPRO-4 & & 19.86 {\scriptsize $\pm 0.18$} & 47.31 {\scriptsize $\pm 0.04$} & 38.93 {\scriptsize $\pm 0.22$} & 46.57 {\scriptsize $\pm 0.08$} & 39.57 {\scriptsize $\pm 0.17$} & 48.84 {\scriptsize $\pm 0.06$} \\
HYPRO-8 & & 19.25 {\scriptsize $\pm 0.30$} & 52.61 {\scriptsize $\pm 0.20$} & 37.30 {\scriptsize $\pm 0.48$} & 53.10 {\scriptsize $\pm 0.17$} & 39.37 {\scriptsize $\pm 0.30$} & 54.14 {\scriptsize $\pm 0.22$} \\
HYPRO-16 & & 18.90 {\scriptsize $\pm 0.34$} & 62.36 {\scriptsize $\pm 0.27$} & 37.08 {\scriptsize $\pm 0.39$} & 65.63 {\scriptsize $\pm 0.44$} & 38.99 {\scriptsize $\pm 0.16$} & 64.82 {\scriptsize $\pm 0.34$} \\
HYPRO-32 & & 18.81 {\scriptsize $\pm 0.16$} & 81.30 {\scriptsize $\pm 0.23$} & 36.96 {\scriptsize $\pm 0.19$} & 89.11 {\scriptsize $\pm 0.71$} & 38.84 {\scriptsize $\pm 0.20$} & 83.05 {\scriptsize $\pm 0.45$} \\
DTPP & 400 & \textbf{15.00} {\scriptsize $\pm 0.30$} & \textbf{0.01} {\scriptsize $\pm 0.00$} & \textbf{28.83} {\scriptsize $\pm 0.26$} & \textbf{0.17} {\scriptsize $\pm 0.01$} & \textbf{36.75} {\scriptsize $\pm 0.40$} & \textbf{0.03} {\scriptsize $\pm 0.00$} \\
\midrule
Speedup & & \multicolumn{2}{c}{$8,130 \times$} & \multicolumn{2}{c}{ $524.2 \times$} & \multicolumn{2}{c}{$2768.3 \times$} \\
\bottomrule
\end{tabular}
\end{sc}
\end{small}
\end{center}

\end{table*}

Apart from the predictive performance, we investigated the time required for the two methods to generate all the predicted sequences of the held-out dataset. Since HYPRO's inference time is heavily relied on the thinning algorithm and a hyperparameter that indicates the number of proposal sequences (denoted as $M$ in~\cite{xue2022hypro}), we conducted an ablation study for a varied number of proposals to investigate the inference time and performance of HYPRO against DTPP. For HYPRO's inference time, apart from the prediction time, we included the time required to generate the noise sequences so the energy function can be trained on. The inclusion of this time is justified by the importance the energy function has as a component of the framework, and it can be seen as a necessary pre-inference step. However, for completeness, we compute only the prediction time of HYPRO and report it in Table~\ref{table:multi_step_pred_time_inference} of the Appendix.

The results are presented in Table~\ref{table:multi_step_pred_otd_vs_time} where we measure the performance using the average OTD; a similar table for RMSE$^\ast$ is in Appendix~\ref{sec:extra_results_long_horizon}. We see that our parallelizable framework takes advantage of modern GPU hardware and performs inference in a few seconds. Instead, the thinning algorithm constitutes HYPRO extremely slow and impractical for inference on large datasets. In some cases like the Taxi dataset, HYPRO needs $8,130 \times$ more time than DTPP to perform inference. Moreover, DTPP attains better performance across all datasets even for a larger number of proposals in HYPRO. These results verify our assumption about the robustness of the mixture model to predict accurately the next time; they also highlight the inaccurate predictions and computational burden of the thinning algorithm.

%% file: discussion.tex
\section{Discussion}\label{sec:discussion}
We have presented DTPP, a Transformer-based probabilistic model for continuous-time event sequences. The model has been derived using the decomposability of the likelihood of a MTPP in terms of its CPDF and CPMF. We have used a mixture of log-normals and a Transformer architecture to model CPDF and CPMF, respectively. Our model satisfies some desirable properties compared to previous works that tried to model the CIF such as closed-form computation of the log-likelihood and inference without resorting to the thinning algorithm. Extensive experiments on the standard task of next-event prediction showed that our method outperformed all state-of-the-art autoregressive models, The results also reveal a more robust performance of the methods that do not require the thinning algorithm to generate event sequences over those they do. Finally, we have tested our model on the challenging task of long-horizon prediction of event sequences. Although our model has not been designed for this task, it outperformed the state-of-the-art baseline HYPRO which is also based on the thinning algorithm. This performance for DTPP was achieved in orders of magnitude faster than HYPRO. 

\paragraph{Limitations and future work.} 
The main limitation of the model stems from the modeling of $p^\ast$ using a deep learning architecture which is usually data-hungry and thus requires large amount of data to learn the model's parameters. For this reason, the model might be unsuitable for data-scarce regimes since it could be prone to overfit. Regarding future work, the limitations of the thinning algorithm revealed by the experiments raise many interesting questions on how can we improve this pathology for the CIF-based methods so they match the performance of the CPDF-based ones since their representations are equivalent. Another interesting research direction would be the development of a globally normalized model similar to HYPRO for CPDF-based models.

%% file: acknowledgements.tex
\section*{Acknowledgements}
We would like to thank Petros Dellaportas and Lina Gerontogianni for helpful discussions.

%% file: appendix.tex
\newpage
\appendix
\onecolumn

\section{Experimental details}\label{sec:exp_details}

\subsection{Dataset Details}\label{sec:dataset_details}
Summary statistics and characteristics of the datasets used are given in Table~\ref{table:stats_datasets}. A more detailed description is given below:
\begin{itemize}
    \item \textbf{Hawkes1-Synthetic.} This dataset contains synthetic event sequences from a univariate Hawkes process sampled using Tick~\cite{bacry2017tick} whose conditional intensity function is defined by
    $\lambda^\ast (t) = \mu + \sum_{t_i < t} \sum_{j=1}^J \alpha_j \beta_j \exp (-\beta_j (t - t_i))$. We use $J=1, \mu=1, \alpha_1 = 0.8, \beta_1 = 1.0 $. This dataset has been used in \cite{omi2019fully}.
    \item \textbf{Hawkes2-Synthetic.} Same as \textbf{Hawkes1-Synthetic} where the parameters here are set as $J=2, \mu=0.2, \alpha_1 = 0.4, \beta_1 = 1.0, \alpha_2 = 0.4, \beta_2 = 20$.
    \item \textbf{SAHP-Synthetic.} We sample sequences from a randomly initialized SAHP model using the thinning algorithm. The number of layers and the dimension of hidden states are 4 and 32, respectively. The same dataset has been used in~\cite{yang2021transformer}.
    \item \textbf{MIMIC-II.} The Multiparameter Intelligent Monitoring in Intensive Care (MIMIC-II) is a medical dataset of de-identified clinical visit records of intensive care unit patients for seven years. There are records of 650 patients/sequences where each one contains the time of the visit and the diagnosis of this visit; there are K=75 unique diseases. We try to predict the time and the diagnosis of a patient.
    \item \textbf{Amazon.} This dataset includes time-stamped user product reviews behavior from
January, 2008 to October, 2018. Each user has a sequence of review events with each event containing the timestamp and category of the reviewed product, with each category corresponding to an event type. As in \cite{xue2023easytpp}, we use a subset of 5200 most active users with an average sequence length of 70 which is comprised of K = 16 event types.
    \item \textbf{Taxi.} This dataset records the times of taxi pick-up and drop-off events across
the five boroughs of the New York city (Manhattan, Brooklyn, Queens, The Bronx, Staten Island). For each borough, we can have pick-up or drop-off event and thus, there are K = 10 event types in total. As in~\cite{xue2022hypro}, we pick a randomly sampled subset of 2000 drivers where each driver has a sequence.
    \item \textbf{Taobao.} The dataset comes from the 2018 Tianchi Big
Data Competition. It consists of time-stamped behavior records (e.g., browsing, purchasing) of
anonymous users on the online shopping platform Taobao from the 25th of November through the 3rd of December in 2017. The K=17 event types represent a category group (e.g. men’s clothing). The browsing sequences of the most active 2000 users are picked as event sequences. The time unit is 3 hours and the average inter-arrival time is 0.06.
    \item \textbf{StackOverflow-V1.} The data comes from the well-known question-answering website StackOverflow where users are encouraged to answer questions so they can earn badges. There are 22 different types of badges. Each sequence corresponds to a user and each event gives the time and the
type of budge a user has been awarded. This dataset is the one processed and used in~\cite{du2016recurrent}.
    \item \textbf{StackOverflow-V2.} A truncated version of the original \textbf{StackOverflow-V1.}. The time unit is 11 days and the average inter-arrival time is 0.95. The dataset has been used in~\cite{xue2022hypro}
\end{itemize}

\begin{table*}[ht]
\caption{Characteristics of the synthetic and real-world datasets.}
\label{table:stats_datasets}
\begingroup
\setlength{\tabcolsep}{9.5pt} 
\renewcommand{\arraystretch}{0.8}
\begin{small}
\begin{center}
{\sc
\begin{tabular}{l rrrrrrr}
\toprule
Dataset & \multirow{2}{*}{$K$} & \multicolumn{3}{c}{\# of Event Tokens} & \multicolumn{3}{c}{Sequence length}  \\
\cmidrule(lr){3-5}  \cmidrule(lr){6-8} 
 &  & \multicolumn{1}{c}{Train} & \multicolumn{1}{c}{Val} & \multicolumn{1}{c}{Dev} & \multicolumn{1}{c}{Min} & \multicolumn{1}{c}{Mean} & \multicolumn{1}{c}{Max} \\
\midrule
Hawkes1-Synthetic & 1 & 65536 & 16384 & 32768 & 64 & 64 & 64 \\
Hawkes2-Synthetic & 1 & 65536 & 16384 & 32768 & 64 & 64 & 64 \\
SAHP-Synthetic & 10 & 71023 & 10242 & 20237 & 81 & 101 & 121 \\
Mimic-II & 75 & 1930 & 252 & 237 & 2 & 4 & 33 \\
Amazon & 16 & 288377 & 40995 & 84048 & 14 & 44 & 94 \\
Taxi & 10 & 51854 & 7404 & 14820 & 36 & 37 & 38 \\
Taobao & 17 & 75205 & 11737 & 28455 & 32 & 57 & 64 \\
StackOverflow-v1 & 22 & 345116 & 38065 & 97233 & 41 & 72 & 736 \\
StackOverflow-v2 & 22 &  90497 & 25762 & 26518 & 41 & 65 & 101 \\
\bottomrule
\end{tabular}
}
\end{center}
\end{small}
\endgroup
\end{table*}

\subsection{Training Details}\label{sec:training_details}
We use the Adam optimizer~\cite{kingma2014adam} with its default settings to train all the models in Section~\ref{sec:experiments}. We use 200 epochs in total, a batch size of 8 sequences, and we apply early-stopping based on the log-likelihood of the held-out dev set.

Following~\cite{yang2021transformer}, we use a common hyperparameter $D$ to define all dimensionalities of the query, key, and value vectors for the models with attention mechanisms, i.e. THP, SAHP, A-NHP, and DTPP-CPMF. For these methods, we also need to specify the number of layers $L$. We also denote by $D$ the state space of the IFTPP model.

The hyperparameters $D$ and $L$ were fine-tuned for each combination of dataset and model. We grid-search  the two parameters using the search spaces $D \in \{ 4, 8, 16, 32, 64, 128 \}$ and $L \in \{1, 2, 3, 4, 5 \}$. We pick the set of values that achieve the highest log-likelihood on the dev set. 

For IFTPP we use the same search for $D$ as before. Since IFTPP and DTPP (CPDF part) are based on a mixture of log-normals, we need to define the number of components $M$. We fine tune $M$ over the space $M \in \{1, 2, 4, 8, 16 \}$.

VI-DPP has only one hyperparameter that requires tuning. This is the number of cutt-off points $Q$~\cite{panos2023scalable}. As the original paper, we found that $Q=1$ works the best in all of our datasets.

For long-horizon prediction experiments, we chose a common size architecture for the HYPRO / DTPP transformer by setting $D=128$ and $L=2$ since this combination worked well on both A-NHP and
DTPP in all datasets in Section~\ref{sec:next_event_pred}.

All experiments were carried out on the same Linux machine with a dedicated reserved GPU used for acceleration.
 
\subsection{Implementation Details}\label{sec:implementation_details}
For A-NHP, we use the public Github repository at \url{https://github.com/yangalan123/anhp-andtt}. For THP and SAHP, we also use the same repository where the corrected implementations of THP and SAHP are provided.
For IFTPP,  we use the public Github repository at \url{https://github.com/shchur/ifl-tpp} while for VI-DPP we use the code at \url{https://github.com/aresPanos/Interpretable-Point-Processes}. Our code will be made publicly available on a Github repository after the review period.

\begin{table}
\setlength{\tabcolsep}{1.1pt}
\caption{Performance comparison  between our model DTPP and various baselines in terms of next-event prediction on Mimic-II dataset. The root mean squared error (RMSE) measures the error of the predicted time of the next event, while the error rate (ERROR-$\%$)  evaluates the error of the predicted mark given the true time. The results (lower is better) are accompanied by 95\% bootstrap confidence intervals. $^\dagger, ^\triangleleft, ^\triangleright$ denote the CIF-based methods, the CPDF-based methods that use a single model, and the ones using a seperate model, respectively.
}\label{table:single_step_pred_mimic}
\begin{center}
\begin{small}
\begin{sc}
\begin{tabular}{lcr}
\toprule
& \multicolumn{2}{c}{ \textbf{Mimic-II}} \\
\cmidrule(lr){2-3} \textbf{Methods} &  \textbf{RMSE} &  \textbf{Error}  \\
\midrule
THP$^\dagger$ & 1.00 {\tiny $\pm 0.13$} & 15.12 {\tiny $\pm 0.99$}  \\
SAHP$^\dagger$ & 1.49 {\tiny $\pm 0.15$} & 16.39 {\tiny $\pm 5.72$}  \\
A-NHP$^\dagger$ & 1.00 {\tiny $\pm 0.19$} & 15.19 {\tiny $\pm 5.16$} \\
IFTPP$^\triangleleft$ & 0.74 {\tiny $\pm 0.39$} & 14.87 {\tiny $\pm 5.33$}  \\
VI-DPP$^\triangleright$ & 0.95 {\tiny $\pm 0.40$} & 16.96 {\tiny $\pm 6.30$}  \\
DTPP$^\triangleright$ & \textbf{0.72} {\tiny $\pm 0.39$} & \textbf{14.49} {\tiny $\pm 5.86$}  \\
\bottomrule
\end{tabular}
\end{sc}
\end{small}
\end{center}
\end{table}

\section{Extra Experimental Results}

\subsection{Results on Mimic-II}\label{sec:extra_results_mimic}
Table \ref{table:single_step_pred_mimic} presents the results for next-event prediction task on Mimic-II dataset. The results follow similar patterns as in Table \ref{table:single_step_pred}.

\subsection{Results on Synthetic datasets}\label{sec:extra_results_synthetic}
\begin{figure*}[hpbt] 
    \vskip 0.2in
    \begin{center}
     \subfigure[Hawkes-1]{%
        \includegraphics[width=0.47\linewidth, height=0.33\textwidth]{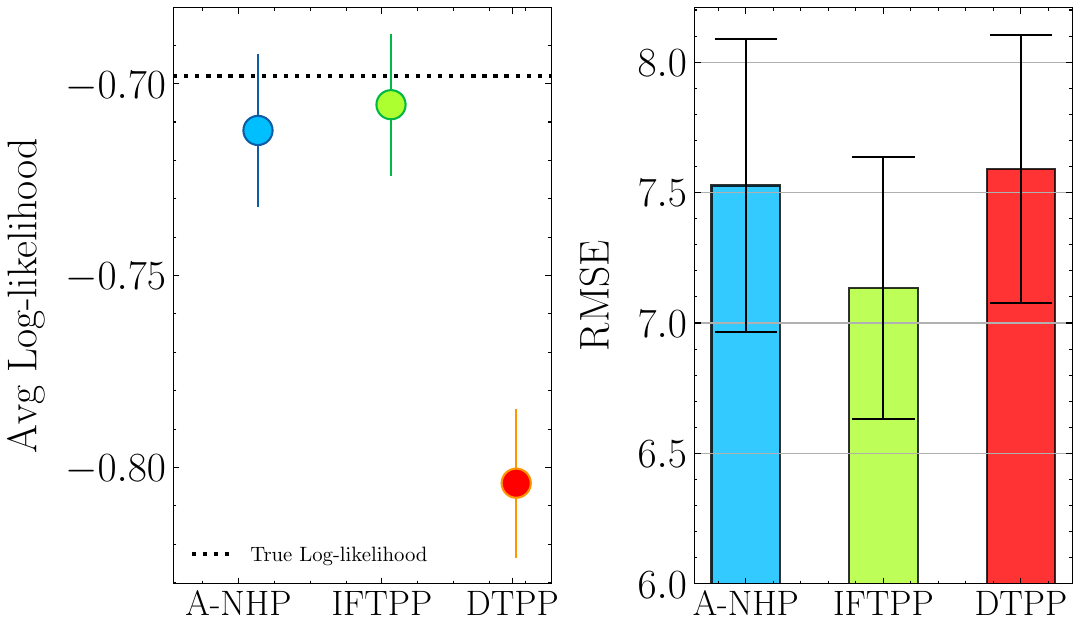}
        \label{fig:hawkes_1l}}
     \subfigure[Hawkes-2]{
        \includegraphics[width=0.47\linewidth, height=0.33\textwidth]{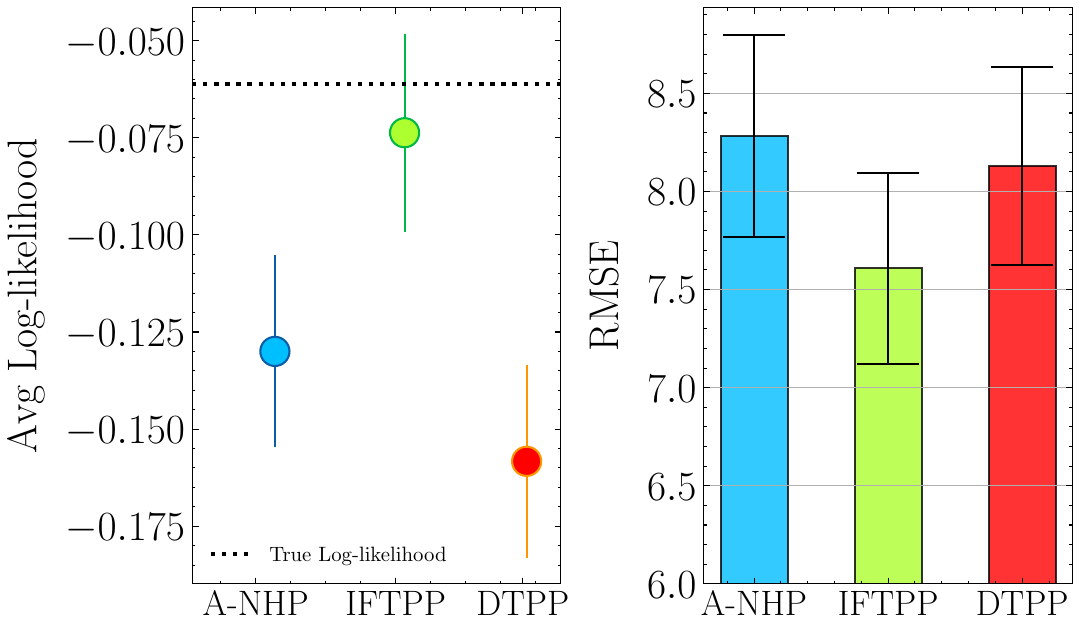}
        \label{fig:hawkes_2}
        }
    \caption{Goodness-of-fit and next-time prediction comparison over the two 1-d synthetic examples generated from a Hawkes process. The reported results are based on the test dataset. The black dotted line represents the true log-likelihood of the data (in nats).}\label{fig:hawkes_synthetic}
    \end{center}
   \vskip -0.2in
\end{figure*}

Figure~\ref{fig:hawkes_synthetic} shows the results for the two synthetic datasets generated by two different one-dimensional Hawkes processes. The DTPP's mixture of log-normals fails to model correctly the two processes since it lacks the flexibility of the neural network that A-NHP and IFTPP are based on. However, when it comes to next-time prediction, DTPP performs on par with A-NHP, showing once more that the thinning algorithm decreases prediction accuracy. IFTPP is the clear winner for these two datasets.

\subsection{Results on Long-Horizon Prediction}\label{sec:extra_results_long_horizon}
We present additional results for the long-horizon prediction in Tables~\ref{table:multi_step_pred_rmse_vs_time} and~\ref{table:multi_step_pred_time_inference}. Table~\ref{table:multi_step_pred_rmse_vs_time} reports the RMSE$^\ast$ where the results show that DTPP outperforms HYPRO, regardless of its number of proposals. The only exception is StackOverflow-v2, where HYPRO achieved a lower score but required at least 1,544 more running time than DTPP.

Finally, we report HYPRO's time without considering the time required to generate noise sequences (Algorithm 1 in~\cite{xue2022hypro}) from the trained auto-regressive model $p_{\text{auto}}$ for training the energy function $E_{\theta}$; see discussion in Section~\ref{sec:long_hor_pred}. Even in this case, our DTPP model is orders of magnitude faster than HYPRO.

\begin{table*}[ht]
\setlength{\tabcolsep}{6.7pt}
\caption{Performance comparison between our model DTPP and HYPRO for the long-horizon prediction task. For HYPRO, we use $\{2, 4, 8, 16, 32 \}$ weighted proposals (Algorithm 2 in~\cite{xue2022hypro}). We report the root mean squared error of the number of tokens for each type of event (RMSE$^{\star}$) and the time (in minutes) required to predict all the long-horizon sequences of the held-out dataset (lower is better). We include error bars for five runs.  
}
\label{table:multi_step_pred_rmse_vs_time}
\vskip 0.15in
\begin{center}
\begin{small}
\begin{sc}
\begin{tabular}{lccccccr}
\toprule
& & \multicolumn{2}{c}{ \textbf{Taxi}} & \multicolumn{2}{c}{ \textbf{Taobao}} & \multicolumn{2}{c}{ \textbf{StackOverflow-V2}} \\
\cmidrule(lr){3-4} \cmidrule(lr){5-6} \cmidrule(lr){7-8} \textbf{Methods} & \textbf{\# Params (K)} &  \textbf{RMSE$^{\star}$} &  \textbf{Time }  &  \textbf{RMSE$^{\star}$} &  \textbf{Time }  &  \textbf{RMSE$^{\star}$} &  \textbf{Time } \\
\midrule
HYPRO-2 & \multirow{5}{*}{850} & 1.39 {\scriptsize $\pm 0.03$} & 44.81 {\scriptsize $\pm 0.01$} & 2.87 {\scriptsize $\pm 0.03$} & 43.53 {\scriptsize $\pm 0.03$} & 1.16 {\scriptsize $\pm 0.01$} & 46.32 {\scriptsize $\pm 0.01$} \\
HYPRO-4 & & 1.34 {\scriptsize $\pm 0.03$} & 47.31 {\scriptsize $\pm 0.04$} & 2.74 {\scriptsize $\pm 0.02$} & 46.57 {\scriptsize $\pm 0.08$} & 1.15 {\scriptsize $\pm 0.01$} & 48.84 {\scriptsize $\pm 0.06$} \\
HYPRO-8 & & 1.27 {\scriptsize $\pm 0.02$} & 52.61 {\scriptsize $\pm 0.20$} & 2.65 {\scriptsize $\pm 0.03$} & 53.10 {\scriptsize $\pm 0.17$} & 1.15 {\scriptsize $\pm 0.02$} & 54.14 {\scriptsize $\pm 0.22$} \\
HYPRO-16 & & 1.24 {\scriptsize $\pm 0.03$} & 62.36 {\scriptsize $\pm 0.27$} & 2.58 {\scriptsize $\pm 0.02$} & 65.63 {\scriptsize $\pm 0.44$} & 1.15 {\scriptsize $\pm 0.02$} & 64.82 {\scriptsize $\pm 0.34$} \\
HYPRO-32 & & 1.20 {\scriptsize $\pm 0.03$} & 81.30 {\scriptsize $\pm 0.23$} & 2.55 {\scriptsize $\pm 0.02$} & 89.11 {\scriptsize $\pm 0.71$} & \textbf{1.14} {\scriptsize $\pm 0.01$} & 83.05 {\scriptsize $\pm 0.45$} \\
DTPP & 400 & \textbf{0.96} {\scriptsize $\pm 0.03$} & \textbf{0.01} {\scriptsize $\pm 0.00$} & \textbf{2.45} {\scriptsize $\pm 0.04$} & \textbf{0.17} {\scriptsize $\pm 0.01$} & 2.02 {\scriptsize $\pm 0.06$} & \textbf{0.03} {\scriptsize $\pm 0.00$} \\
\midrule
Speedup & & \multicolumn{2}{c}{$8,130 \times$} & \multicolumn{2}{c}{ $524.2 \times$} & \multicolumn{2}{c}{$2768.3 \times$} \\
\bottomrule
\end{tabular}
\end{sc}
\end{small}
\end{center}
\vskip -0.1in
\end{table*}

\begin{table*}[ht]
\setlength{\tabcolsep}{6.7pt}
\caption{Time comparison (in minutes) between our model DTPP and HYPRO for the long-horizon prediction task. Unlike Tables~\ref{table:multi_step_pred_otd_vs_time} and~\ref{table:multi_step_pred_rmse_vs_time}, here we only report the prediction time of HYPRO without including the time required to generate noise sequences (Algorithm 1 in~\cite{xue2022hypro}) from the trained auto-regressive model $p_{\text{auto}}$ for training the energy function $E_{\theta}$ . We include error bars based on five runs.  
}
\label{table:multi_step_pred_time_inference}
\vskip 0.15in
\begin{center}
\begin{small}
\begin{sc}
\begin{tabular}{lcccc}
\toprule
\textbf{Methods} & \textbf{\# Params (K)} &  \textbf{Taxi} & \textbf{Taobao} & \textbf{StackOverflow-v2} \\
\midrule
HYPRO-2 & \multirow{5}{*}{850} & 2.41 {\scriptsize $\pm 0.01$} & 3.13 {\scriptsize $\pm 0.03$} & 2.48 {\scriptsize $\pm 0.01$} \\
HYPRO-4 &  & 4.91 {\scriptsize $\pm 0.04$} & 6.17 {\scriptsize $\pm 0.08$} & 5.01 {\scriptsize $\pm 0.06$} \\
HYPRO-8 &  & 10.21 {\scriptsize $\pm 0.20$} & 12.70 {\scriptsize $\pm 0.17$} & 10.30 {\scriptsize $\pm 0.22$} \\
HYPRO-16 &  & 19.96 {\scriptsize $\pm 0.27$} & 25.23 {\scriptsize $\pm 0.44$} & 20.98 {\scriptsize $\pm 0.34$} \\
HYPRO-32 &  & 38.90 {\scriptsize $\pm 0.23$} & 48.70 {\scriptsize $\pm 0.71$} & 39.21 {\scriptsize $\pm 0.44$} \\
DTPP & 400 & \textbf{0.01} {\scriptsize $\pm 0.00$} & \textbf{0.17} {\scriptsize $\pm 0.01$} & \textbf{0.03} {\scriptsize $\pm 0.00$} \\
\midrule
Speedup & & $3,890 \times$ & $286.5 \times$ & $1,307 \times$ \\
\bottomrule
\end{tabular}
\end{sc}
\end{small}
\end{center}
\vskip -0.1in
\end{table*}

\begin{algorithm}
   \caption{Long-Horizon Prediction for Decomposed Transformer Point Processes}\label{alg:long_horizon}
   {\bfseries Input:}~an observed sequence $\Hcal_{T}$ of $N$ events over the interval $(0, T)$ and the number of prediction steps $P > 1$; trained inter-event model $g$ and event types model $p$ \\
   {\bfseries Output:}~predicted sequence $\hat{\Hcal}_P$ of $P$ events \\
   \vspace{-0.5cm}
\begin{algorithmic}[1]
   \STATE {\bfseries function} LHP($\Hcal_{T}$, $P$, $g^\ast, p^\ast$)
   \STATE ~~~~~$\hat{\Hcal}_P \gets \emptyset,~~\hat{\Hcal} \gets \Hcal_{T}$ 
   \STATE ~~~~~{\bfseries for} $p=1$ {\bfseries to} $P$:
   \STATE ~~~~~~~~~~ $\triangleright$~\emph{Predict next-event time and event type}
   \STATE ~~~~~~~~~~ $t_{N + p}  = t_{N + p - 1} + \mathbb{E}^{(k_{N + p - 1})}_g[\tau \mid \hat{\Hcal}]$  
   \STATE ~~~~~~~~~~ $k_{N + p}  =  \argmax_k p(k \mid t_{N+p}, \hat{\Hcal})$  
   \STATE ~~~~~~~~~~ $\triangleright$~\emph{Update predicted sequence and event history}
   \STATE ~~~~~~~~~~ $\hat{\Hcal}_P \gets \hat{\Hcal}_P \cup (t_{N + p}, k_{N + p})$  
   \STATE ~~~~~~~~~~ $\hat{\Hcal} \gets \hat{\Hcal} \cup (t_{N + p}, k_{N + p})$  \\
   \STATE ~~~~~{\bfseries end for}
   \STATE ~~~~~{\bfseries return}~$\hat{\Hcal}_P$
   \STATE  {\bfseries end function}
\end{algorithmic}
\end{algorithm}

\begin{algorithm}[tb]
   \caption{Thinning Algorithm}\label{alg:thinning}
   {\bfseries Input:}~an observed sequence $\Hcal_{T}$ of $N$ events over the interval $(0, T)$ and the number of prediction steps $P > 1$; intensity function $\lambda_{k} $ \\
   {\bfseries Output:}~predicted sequence $\hat{\Hcal}_P$ of $P$ events \\
   \vspace{-0.5cm}
\begin{algorithmic}[1]
   \STATE {\bfseries function} Thinning($\Hcal_{T}$, $P$, $\lambda_{k} $)
   \STATE ~~~~~$\hat{\Hcal}_P \gets \emptyset,~~\hat{\Hcal} \gets \Hcal_{T},~~p \gets 0$,~~$\hat{t} \gets T$ 
   \STATE ~~~~~{\bfseries for} $p=1$ {\bfseries to} $P$:
   \STATE ~~~~~~~~~~ $\lambda_{\text{max}} = \max_{t \in (\hat{t}, \infty)} \sum_{k=1}^K \lambda_{k} (t \mid \hat{\Hcal})$~~~~~~~~~~~~~~~~~~~~~~~~~~~~~~~~~~~~~~~~~~~~~~~~~~~~~~~~~~~~~~~~~~~~~~ $\triangleright$~\emph{Compute upper bound $ \lambda_{\text{max}}$}  
   \STATE ~~~~~~~~~~{\bfseries repeat}:
   \STATE ~~~~~~~~~~$\tau \sim \text{Exp}(\lambda_{\text{max}})$
   \STATE ~~~~~~~~~~$\hat{t} \gets \hat{t} + \tau$
   \STATE ~~~~~~~~~~{\bfseries until} $u  \le  \sum_{k=1}^K \lambda_{k} (\hat{t} \mid \hat{\Hcal}) / \lambda_{\text{max}}$ ~~~~~~~~$\triangleright$~\emph{Accept proposed occurrence time with probability $\sum_{k=1}^K \lambda_{k} (\hat{t} \mid \hat{\Hcal}) / \lambda_{\text{max}}$}
   \STATE ~~~~~~~~~~~$\hat{k} \sim \text{Cat}(p_1, \ldots, p_K)$ where $p_k = \lambda_{k} (\hat{t} \mid \hat{\Hcal}) / \sum_{k=1}^K \lambda_{k} (\hat{t} \mid \hat{\Hcal})$ ~~~~~~~~~~~~~~~~~~~~~ $\triangleright$~\emph{Sample event type $\hat{k} \in \{1, \ldots, K \}$ } 
   \STATE ~~~~~~~~~ $t_{N + p} \gets \hat{t},~~ k_{N + p} \gets \hat{k}$
   \STATE ~~~~~~~~~~ $\hat{\Hcal}_P \gets \hat{\Hcal}_P \cup (t_{N + p}, k_{N + p}),~~~\hat{\Hcal} \gets \hat{\Hcal} \cup (t_{N + p}, k_{N + p})$  ~~~~~~~~~~~~~$\triangleright$~\emph{Update predicted sequence and event history}
   \STATE ~~~~~{\bfseries end for}
   \STATE ~~~~~{\bfseries return}~$\hat{\Hcal}_P$
   \STATE  {\bfseries end function}
\end{algorithmic}
\end{algorithm}

%% file: neurips_checklist.tex
\newpage
\appendix
\onecolumn

\section*{NeurIPS Paper Checklist}

The checklist is designed to encourage best practices for responsible machine learning research, addressing issues of reproducibility, transparency, research ethics, and societal impact. Do not remove the checklist: {\bf The papers not including the checklist will be desk rejected.} The checklist should follow the references and follow the (optional) supplemental material.  The checklist does NOT count towards the page
limit. 

Please read the checklist guidelines carefully for information on how to answer these questions. For each question in the checklist:
\begin{itemize}
    \item You should answer \answerYes{}, \answerNo{}, or \answerNA{}.
    \item \answerNA{} means either that the question is Not Applicable for that particular paper or the relevant information is Not Available.
    \item Please provide a short (1–2 sentence) justification right after your answer (even for NA). 
\end{itemize}

{\bf The checklist answers are an integral part of your paper submission.} They are visible to the reviewers, area chairs, senior area chairs, and ethics reviewers. You will be asked to also include it (after eventual revisions) with the final version of your paper, and its final version will be published with the paper.

The reviewers of your paper will be asked to use the checklist as one of the factors in their evaluation. While "\answerYes{}" is generally preferable to "\answerNo{}", it is perfectly acceptable to answer "\answerNo{}" provided a proper justification is given (e.g., "error bars are not reported because it would be too computationally expensive" or "we were unable to find the license for the dataset we used"). In general, answering "\answerNo{}" or "\answerNA{}" is not grounds for rejection. While the questions are phrased in a binary way, we acknowledge that the true answer is often more nuanced, so please just use your best judgment and write a justification to elaborate. All supporting evidence can appear either in the main paper or the supplemental material, provided in appendix. If you answer \answerYes{} to a question, in the justification please point to the section(s) where related material for the question can be found.

IMPORTANT, please:
\begin{itemize}
    \item {\bf Delete this instruction block, but keep the section heading ``NeurIPS paper checklist"},
    \item  {\bf Keep the checklist subsection headings, questions/answers and guidelines below.}
    \item {\bf Do not modify the questions and only use the provided macros for your answers}.
\end{itemize}


\begin{enumerate}

\item {\bf Claims}
    \item[] Question: Do the main claims made in the abstract and introduction accurately reflect the paper's contributions and scope?
    \item[] Answer: \answerYes{} 
    \item[] Justification: See Section \ref{sec:introduction}

\item {\bf Limitations}
    \item[] Question: Does the paper discuss the limitations of the work performed by the authors?
    \item[] Answer: \answerYes{} 
    \item[] Justification: See section \ref{sec:discussion}

\item {\bf Theory Assumptions and Proofs}
    \item[] Question: For each theoretical result, does the paper provide the full set of assumptions and a complete (and correct) proof?
    \item[] Answer: \answerNA{} 
    \item[] Justification: -

    \item {\bf Experimental Result Reproducibility}
    \item[] Question: Does the paper fully disclose all the information needed to reproduce the main experimental results of the paper to the extent that it affects the main claims and/or conclusions of the paper (regardless of whether the code and data are provided or not)?
    \item[] Answer: \answerYes{} 
    \item[] Justification: See Sections \ref{sec:experiments} and \ref{sec:implementation_details}

\item {\bf Open access to data and code}
    \item[] Question: Does the paper provide open access to the data and code, with sufficient instructions to faithfully reproduce the main experimental results, as described in supplemental material?
    \item[] Answer: \answerYes{} 
    \item[] Justification: See Sections \ref{sec:experiments} and \ref{sec:implementation_details}

\item {\bf Experimental Setting/Details}
    \item[] Question: Does the paper specify all the training and test details (e.g., data splits, hyperparameters, how they were chosen, type of optimizer, etc.) necessary to understand the results?
    \item[] Answer: \answerYes{} 
    \item[] Justification: See Sections \ref{sec:experiments} and \ref{sec:implementation_details}

\item {\bf Experiment Statistical Significance}
    \item[] Question: Does the paper report error bars suitably and correctly defined or other appropriate information about the statistical significance of the experiments?
    \item[] Answer: \answerYes{} 
    \item[] Justification: \justificationTODO{}
    \item[] Guidelines: See Section \ref{sec:experiments}
    \begin{itemize}
        \item The answer NA means that the paper does not include experiments.
        \item The authors should answer "Yes" if the results are accompanied by error bars, confidence intervals, or statistical significance tests, at least for the experiments that support the main claims of the paper.
        \item The factors of variability that the error bars are capturing should be clearly stated (for example, train/test split, initialization, random drawing of some parameter, or overall run with given experimental conditions).
        \item The method for calculating the error bars should be explained (closed form formula, call to a library function, bootstrap, etc.)
        \item The assumptions made should be given (e.g., Normally distributed errors).
        \item It should be clear whether the error bar is the standard deviation or the standard error of the mean.
        \item It is OK to report 1-sigma error bars, but one should state it. The authors should preferably report a 2-sigma error bar than state that they have a 96\% CI, if the hypothesis of Normality of errors is not verified.
        \item For asymmetric distributions, the authors should be careful not to show in tables or figures symmetric error bars that would yield results that are out of range (e.g. negative error rates).
        \item If error bars are reported in tables or plots, The authors should explain in the text how they were calculated and reference the corresponding figures or tables in the text.
    \end{itemize}

\item {\bf Experiments Compute Resources}
    \item[] Question: For each experiment, does the paper provide sufficient information on the computer resources (type of compute workers, memory, time of execution) needed to reproduce the experiments?
    \item[] Answer: \answerYes{} 
    \item[] Justification: See Section \ref{sec:exp_details}
    
\item {\bf Code Of Ethics}
    \item[] Question: Does the research conducted in the paper conform, in every respect, with the NeurIPS Code of Ethics \url{https://neurips.cc/public/EthicsGuidelines}?
    \item[] Answer: \answerYes{} 
    \item[] Justification: 

\item {\bf Broader Impacts}
    \item[] Question: Does the paper discuss both potential positive societal impacts and negative societal impacts of the work performed?
    \item[] Answer: \answerNo{} 
    \item[] Justification: This paper presents work whose goal is to advance the field of Machine Learning. There are many potential societal consequences of our work, none which we feel must be specifically highlighted here.
    \item[] Guidelines: This work is concerned with modeling point processes and this 
    \begin{itemize}
        \item The answer NA means that there is no societal impact of the work performed.
        \item If the authors answer NA or No, they should explain why their work has no societal impact or why the paper does not address societal impact.
        \item Examples of negative societal impacts include potential malicious or unintended uses (e.g., disinformation, generating fake profiles, surveillance), fairness considerations (e.g., deployment of technologies that could make decisions that unfairly impact specific groups), privacy considerations, and security considerations.
        \item The conference expects that many papers will be foundational research and not tied to particular applications, let alone deployments. However, if there is a direct path to any negative applications, the authors should point it out. For example, it is legitimate to point out that an improvement in the quality of generative models could be used to generate deepfakes for disinformation. On the other hand, it is not needed to point out that a generic algorithm for optimizing neural networks could enable people to train models that generate Deepfakes faster.
        \item The authors should consider possible harms that could arise when the technology is being used as intended and functioning correctly, harms that could arise when the technology is being used as intended but gives incorrect results, and harms following from (intentional or unintentional) misuse of the technology.
        \item If there are negative societal impacts, the authors could also discuss possible mitigation strategies (e.g., gated release of models, providing defenses in addition to attacks, mechanisms for monitoring misuse, mechanisms to monitor how a system learns from feedback over time, improving the efficiency and accessibility of ML).
    \end{itemize}
    
\item {\bf Safeguards}
    \item[] Question: Does the paper describe safeguards that have been put in place for responsible release of data or models that have a high risk for misuse (e.g., pretrained language models, image generators, or scraped datasets)?
    \item[] Answer: \answerNA{} 
    \item[] Justification:

\item {\bf Licenses for existing assets}
    \item[] Question: Are the creators or original owners of assets (e.g., code, data, models), used in the paper, properly credited and are the license and terms of use explicitly mentioned and properly respected?
    \item[] Answer: \answerYes{} 
    \item[] Justification: See Sections \ref{sec:experiments} and \ref{sec:dataset_details}

\item {\bf New Assets}
    \item[] Question: Are new assets introduced in the paper well documented and is the documentation provided alongside the assets?
    \item[] Answer: \answerNA{} 
    \item[] Justification: 

\item {\bf Crowdsourcing and Research with Human Subjects}
    \item[] Question: For crowdsourcing experiments and research with human subjects, does the paper include the full text of instructions given to participants and screenshots, if applicable, as well as details about compensation (if any)? 
    \item[] Answer: \answerNA{} 
    \item[] Justification: 

\item {\bf Institutional Review Board (IRB) Approvals or Equivalent for Research with Human Subjects}
    \item[] Question: Does the paper describe potential risks incurred by study participants, whether such risks were disclosed to the subjects, and whether Institutional Review Board (IRB) approvals (or an equivalent approval/review based on the requirements of your country or institution) were obtained?
    \item[] Answer: \answerNA{} 
    \item[] Justification: 

\end{enumerate}